# A reformulation of collision avoidance algorithm based on artificial potential fields for fixed-wing UAVs in a dynamic environment

**Astik Srivastava\*. PB Sujit.\*\***

*\*Delhi Technological University, New Delhi, 110042
India (Tel: 9311784289; e-mail: astiksrivastava_ep20a18_49@dtu.ac.in).*

*\*\*Indian Institute of Science, Education, and Research, Bhopal, Madhya Pradesh, India (e-mail: sujit@iiserb.ac.in)*

**Abstract**: As mini UAVs become increasingly useful in a civilian work domain, the need for a method for them to operate safely in a cluttered environment is growing, especially for fixed-wing UAVs as they are incapable of following the stop-decide-execute methodology. This paper presents preliminary research to design a reactive collision avoidance algorithm based on the improved definition of the repulsive forces used in the Artificial potential field algorithms to allow feasible and safe navigation of fixed-wing UAVs in cluttered, dynamic environments. We present simulation results of the improved definition in multiple scenarios, and we have also discussed possible future studies to improve upon these results.

## 1. INTRODUCTION

Unmanned Aerial Vehicles (UAVs) have been a vital asset for most militaries worldwide for the past two decades. In recent years, they have also started attracting significant attention in the civilian domain due to their potential for cost-effectively solving problems like organ transportation and agricultural mapping.

As these UAVs become a more integral part of the day-to-day lives of an average person, the airspace, especially in urban areas, will become cluttered and difficult to manage. Cluttered airspace also means that the probability of inter-vehicular collisions and collisions of UAVs with other objects in the environment will be high.

The Artificial Potential Field Approach (APFA) was first proposed by *Khatib et al. (1986)*. It is one of the most promising approaches for solving the collision avoidance problem due to its mathematical simplicity and computational efficiency, but the algorithm has limitations.

The first drawback is the lack of a guarantee that the path taken by the UAV is optimum. The second and, in most cases, most catastrophic drawback is the tendency of the agent to get stuck in a local minimum, which is generally the case in environments with complex geometry or clutter.

In this paper, we propose our version of the potential field algorithm that has proven to minimise the probability of encountering a local minimum and is also inclined towards providing a more optimum path. We achieved this by reformulating the repulsive potential function used in prior works. The novel potential function introduced in this paper is based on the **Cauchy distribution.** The contour of the repulsive potential function in the x-y plane is of the shape of an ellipse, contrary to the circular contours seen in typical APFA (Artificial Potential Field Approaches), whose major axes always point in the direction of the relative velocity of the oncoming agent (UAV in this case) w.r.t the obstacle. We suggest a strategy to tune the gains of the potential function based on the likelihood of collision with an obstacle. The main advantage that this reformulation has demonstrated is that it generates feasible commands for a fixed-wing UAV while minimising the deviation from the desired path. To the best of our knowledge, only work in robotic literature that uses a similar concept with a different mathematical formulation is done by *TJ Stastny et al*.

The current work is limited to solving the obstacle avoidance problem in a two-dimensional environment, although the algorithm can be extended to a 3D environment with some modifications.

## 2. Algorithm Formulation

### 2.1 Problem Formulation

The algorithm proposed in this paper aims at performing reactive collision avoidance with minimal deviation from the path taken by the UAV. Since most mini UAVs lack control surfaces to decelerate quickly, the airspeed of the UAV is assumed to be constant. Also, the current work aims at performing obstacle avoidance and trajectory following in a 2-dimensional environment. Hence the altitude is also kept constant.

The algorithm is primarily designed for the safe operation of fixed-wing UAVs, as generally, they move at high velocities and have limited manoeuvrability. Obstacle avoidance is achieved by constructing a potential field around an obstacle with an elliptical contour in the x-y plane, as shown (Fig. 1).

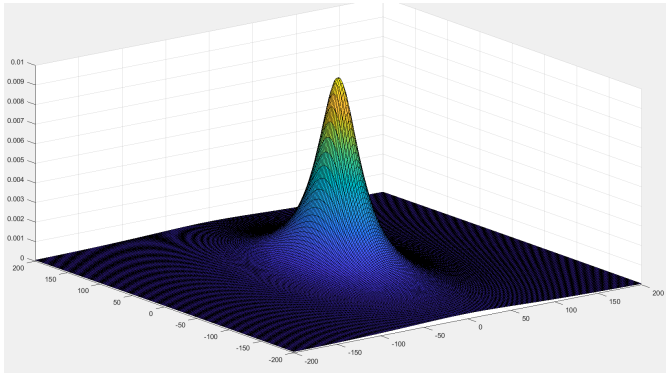

(a)

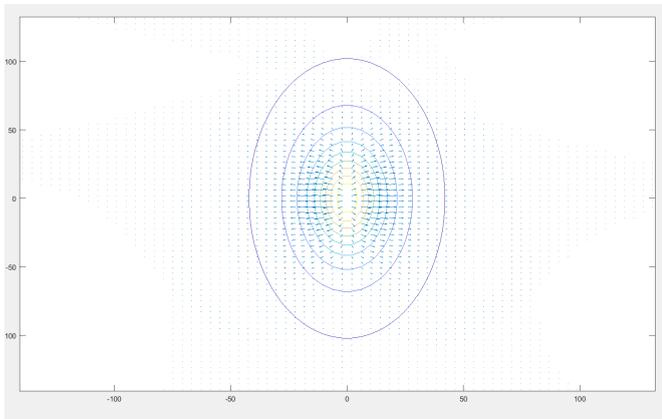

(b)

Figure 1. (a) Potential field generated for an obstacle placed at the origin, with a = 14 and b = 34. (b) gradient and contour plot of the repulsive potential function.

The major axis of the contours is parallel to the relative velocity between the obstacle and the UAV. The parallelism allows the UAV to start taking small manoeuvres when it is far from the obstacle rather than making sharp manoeuvres when it is near it. This guidance strategy reduces the perpendicular deviation of the UAV from the intended path, providing some optimality and ensuring that the manoeuvres commanded to the UAV are within its feasible range.

### 2.2 Guidance model of the UAV

The equations of motion for the fixed-wing UAV used are shown below(1):

$$\dot{x} = V_g * cos(\psi)$$

$$\dot{y} = V_g * sin(\psi) \qquad (1)$$

$$\dot{\psi} = \frac{g}{V} * tan(\phi)$$

This model is based on the assumption that the UAV is flying in a coordinated-turn condition, i.e., side slip angle is 0 and has been implemented using MATLAB's inbuilt Fixed-wing guidance model. More intricate control schemes like MPC, LQG controllers, etc., can be implemented to track the heading commands given by the obstacle avoidance algorithm (see *Stastny et al.(2013))*, but they are beyond the scope of this paper.

### 2.3 Potential field-based obstacle avoidance

Potential field algorithm works by using the gradients of a scalar potential field to guide the agent towards the desired point or along a desired path. This method is analogous to the behaviour of charges in an electric field. The agent and obstacles are analogous to like-charges, whereas the agent and the goal are analogous to opposite charges. The resulting forces then guide the UAV away from obstacles and towards the goal.

The net potential and net force experienced by the agent can be represented as shown in (2):

$$U_{net} = \sum_{i=1}^{n} U(i)_{rep} + U_{att} \qquad (2)$$

$$F_{net} = \sum_{i=1}^{n} \nabla U(i)_{rep} + \nabla U_{att}$$

The acceleration itself is not used to guide the UAV. Instead, the heading is changed to point towards the direction of the resultant force while keeping the speed constant (3).

$$\psi = tan^{-1}(\frac{F_y}{F_x}) \qquad (3)$$

## 2.4 Attractive potential fields

The attractive potential field to guide the UAV towards a point is defined as (4)

$$U_{att} = k_{att} * ((x - x_{wp})^2 + (y - y_{wp})^2) \quad (4)$$

where $k_{att}$ is a tunable parameter, $(x_{wp}, y_{wp})$ are the coordinates of the target waypoint, and $(x,y)$ is the position of the agent. The force due to this potential field is given by

$$F_{att} = 2 * k_{att} * [(x - x_{wp})\hat{i} + (y - y_{wp})\hat{j}] \quad (5)$$

Although this formulation is effective in guiding the UAV towards a point, there is no fixed path that the UAV takes to reach that point when using this force. The UAV can be constrained to follow a straight/curved path from point A to point B or follow an orbit around a point with a given radius based on vector fields given by *Nelson et al. (2007)*.

## 2.5 Repulsive potential field

The potential function around the obstacle is designed to allow smooth manoeuvring. It also minimises the deviation from the desired path while ensuring that the minimum distance between the obstacle and UAV is always greater than the *minimum-allowable separation*. A 3D Cauchy distribution with an elliptical base is used as a template for the potential function(fig 1.). The function is subject to rotation transformation about the z-axis so that the major axis of the potential field is parallel to the relative velocity vector of the incoming UAV w.r.t obstacle.(fig 2).

The potential function is given as follows (7):

$$U_{rep} = \sum_{i=0}^{n} \frac{K_{rep(i)}}{1 + (\frac{X}{a(i)})^2 + (\frac{Y}{b(i)})^2} \quad (7)$$

where:

$$X = (x_{pos} - x_{obs(i)}) * cos(\theta_i) + (y_{pos} - y_{obs(i)}) * sin(\theta_i)$$

$$Y = (x_{pos} - x_{obs(i)}) * sin(\theta_i) - (y_{pos} - y_{obs(i)}) * cos(\theta_i)$$

$$\theta_i = tan^{-1}(\frac{V_{y(i)}}{V_{x(i)}})$$

here, $k_{rep(i)}$ is the amplitude of the function, $(x_{pos}, y_{pos})$ is the position of the UAV, $(x_{obs(i)}, y_{obs(i)})$ is the position of the $i^{th}$ obstacle, and $\theta_i$ is the inclination of the relative velocity of the $i^{th}$ obstacle w.r.t UAV, from the x-axis. In this formulation, only the length of the semi-minor axis of the ellipse is fixed and depends on the dimensions of the obstacle (or can be set to the desired minimum-safe-distance required between the UAV and the obstacle). Parameters $a$ and $k_{rep}$ dynamically change according to the scenario, as will be demonstrated.

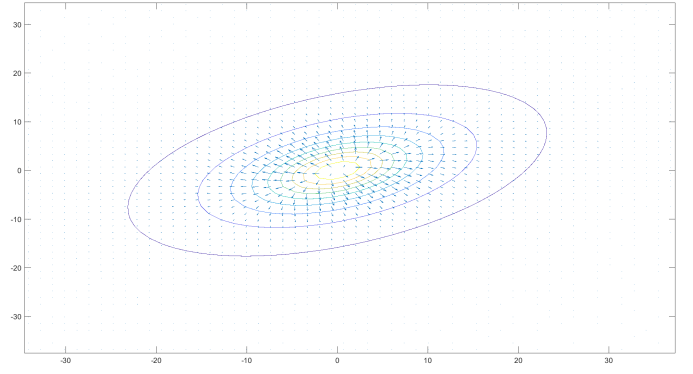

Figure 2. Potential field inclined at an angle of π/6 with a = 8 and b = 6.

We obtain the gradient of the potential field by multiplying the gradients of the non-rotated function (with major and minor axes along the x-axis and y-axis, respectively) with the transformation matrix given by

$$R = \begin{bmatrix} \cos\theta & -\sin\theta \\ \sin\theta & \cos\theta \end{bmatrix}$$

But, the final vectors which give the heading to the UAV are slightly different from the mathematical gradient of the potential field.

The result can be expressed as (8):

$$F_{r(i)} = F_{rep(i)}\hat{i} + F_{rep(i)}\hat{j} \quad (8)$$

$$F_{r(i)}\hat{i} = -2 * k_{rep(i)}(u_{(i)})^{-2} * (\frac{X}{a(i)^2} * cos(\theta(i)) + \frac{Y}{b(i)^2} * sin(\theta(i)))$$

$$F_{r(i)}\hat{j} = -2 * k_{rep(i)}(u_{(i)})^{-2} * (\frac{X}{a(i)^2} * sin(\theta(i)) + \frac{Y}{b(i)^2} * cos(\theta(i)))$$

where $\hat{i}$ and $\hat{j}$ are unit vectors along x and y-axis respectively. The coefficient $u_{(i)}$ is defined as

$$u_{(i)} = 1 + \Gamma * ((\frac{X}{a(i)})^2 + (\frac{Y}{b(i)})^2) \quad (9)$$

The constant Γ is introduced to shift the maxima of the magnitude of force $F_{rep}$, and **X** and **Y** are defined in (7) This constant is fixed to **0.34** so that the maxima lie on the minor axis, at a distance of b from the obstacle.

Parameters $a_i$, and $k_{rep(i)}$ determine how far from the obstacle will the UAV start taking the evasive manoeuvres and how far will the UAV be deflected from its intended path, respectively. For the algorithm to produce desirable results, these two parameters vary depending on the likelihood of collision with the obstacle. This is done as follows:

We define $\eta_i$ as the angle between the position vector from UAV to obstacle(i) and the relative velocity vector, then

$$\xi_i = cos(\eta_i) = \frac{(x_{obs(i)}-x_{UAV})*cos(\theta_{(i)})+(y_{obs(i)}-y_{UAV})*sin(\theta_{(i)})}{\sqrt{(x_{obs(i)}-x_{UAV})^2+(y_{obs(i)}-y_{UAV})^2}}$$

$$k_{rep(i)} = k_{rep0} * (sin(\frac{\pi*\xi_i}{2}) + 1) \qquad (10)$$

$$a_{(i)} = b_{(i)}(1.734 + \xi_i)$$

Variation of $k_{rep(i)}$ has been shown in fig.3. As you can see, when the collision is head-on, i.e. η = 0, then the amplitude of repulsive potential is maximum, and then decreases as the angle increases.

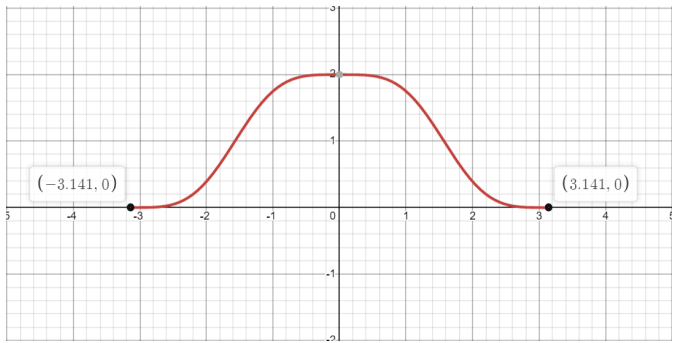

Figure 3. variation of $k_{rep}$ (y-axis) with angle η (x-axis, between -π to π)

The coefficient $k_{rep0}$ has been tuned heuristically for head-on collisions.

This methodology of adapting the eccentricity of contours and amplitude works well in cluttered environments. More complex adaptive algorithms based on the likelihood of collision derived using the collision cone algorithm (see *Mujumdar A (2011)*) can be used.

### 3. Simulation results and discussion

To test the algorithm's validity, multiple scenarios were constructed with both static and dynamic obstacles.

*3.1 Simulation 1*

The first simulation is a set of scenarios designed to test the capability of the proposed potential field algorithm in a head-on collision and when passing a narrow space between two obstacles, wherein traditional APFA tends to get stuck in local minima.

The case of Goal-Not-Reachable, where the agent is unable to reach the goal as it is close to an obstacle, is not considered as that situation is unlikely to be encountered in a real-life.

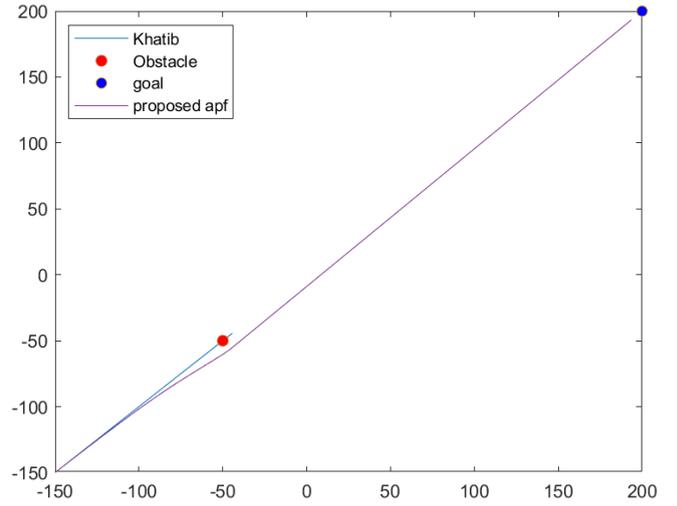

(a)

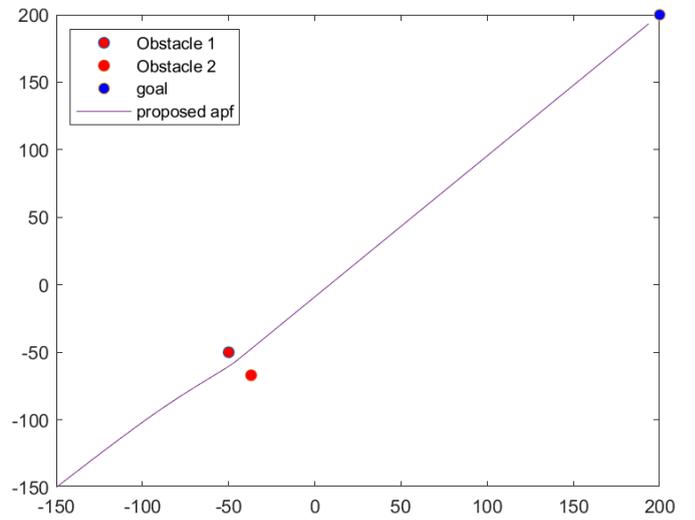

(b)

Figure 4. (a) Scenario showing the path taken by UAV with traditional APFA and with proposed APFA for a head-on collision. Path taken by the UAV in the presence of two obstacles closely spaced.

*3.2 Simulation 2*

The second simulation demonstrates dynamic obstacle avoidance in a multi-agent environment. Each agent independently tries to avoid the other agents and reach its goal. The goal for each UAV is the initial position of the UAV diagonally opposite, causing all agents to converge in the center. The results of this scenario are given below.

## 3.2 Simulation 3

This simulation scenario has been simulated using full lateral kinematics of the UAV as shown in (1). MATLAB's UAV Toolbox has been used to design the scenario and incorporate equations (1) in the system. The UAV attempted to reach the goal while avoiding three buildings and one UAV moving with a constant velocity. The path taken by UAVs and their roll angle response are shown in figure 5. The parameters of the scenario are shown in Table 2.

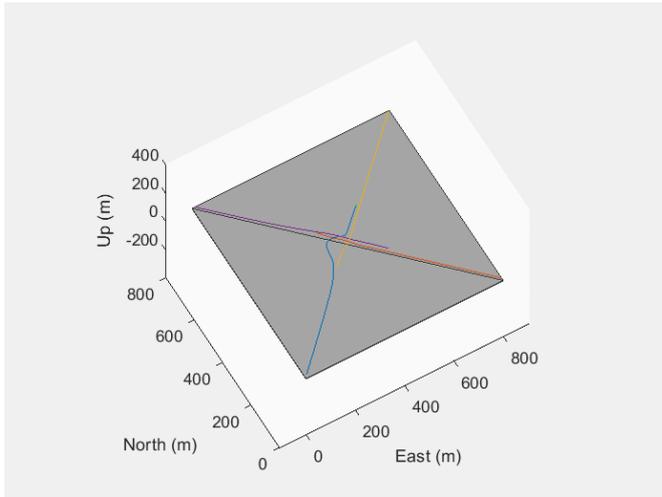

(a)

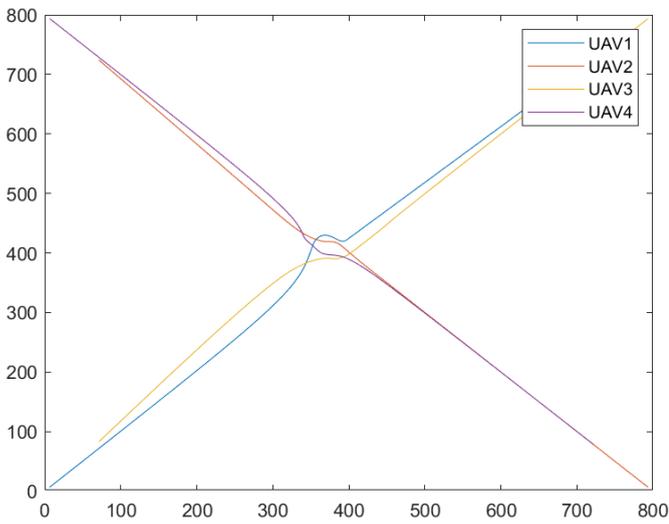

(b)

Figure 5. Path taken by 4 agents to collaboratively avoid collision

**Table 1 Potential function parameters**

| Parameters | Value | Units |
|---|---|---|
| Start position | UAV1 = [0, 0]<br>UAV2 = [800,0]<br>UAV3 = [800,800]<br>UAV4 = [0,800] | meters |
| Goal position | UAV1 = [800,800]<br>UAV2 = [0,800]<br>UAV3 = [0,0]<br>UAV4 = [800,0] | meters |
| $k_{att}$ | 0.008 | - |
| $K_{rep0}$ | 30 | - |
| b | [30, 30, 30, 30] | meters |

**Table 2 Simulation parameters**

| Parameters | Value | Units |
|---|---|---|
| Start position | (0, 0) | meters |
| Goal position | (800, 800) | meters |
| $k_{att}$ | 0.008 | - |
| $K_{rep0}$ | 30 | - |
| b for static obstacles | [11; 6; 9] | meters |
| b for dynamic obstacles | [20] | meters |
| Command Airspeed of UAV (V) | 15 | meters/sec |
| positions of static obstacles. | [500, 550; 450, 500; 250, 250] | meters |
| $V_{ob2}$ (velocity of obstacle 1) | (0,-10) | meters/sec |
| Co-ordinate frame convention | East-North-UP | |

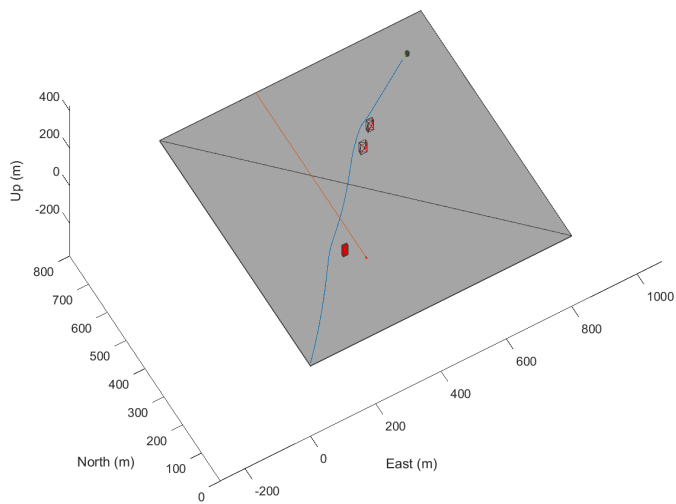

(a)

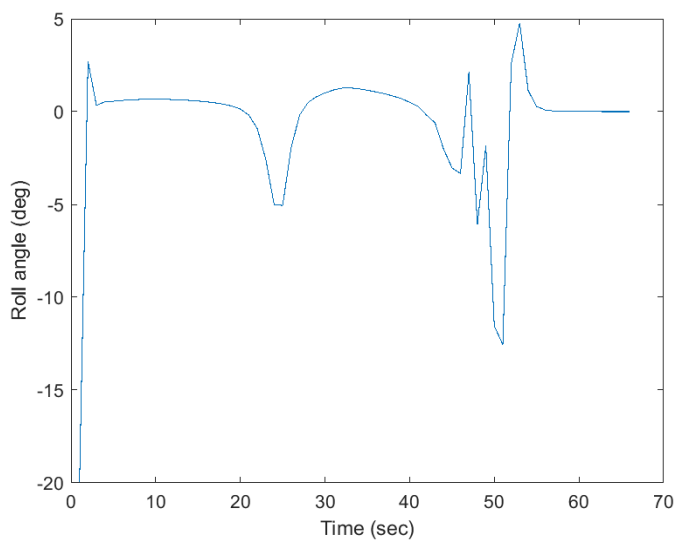

(b)

Figure 6. (a) Scenario with the red path being that of obstacle UAV, and blue being that of the controlled UAV. (b) The roll angle response of the UAV.

## 6. CONCLUSIONS

This paper attempts to develop a new expression for the repulsive force used in the APF approach to guide UAVs. The modified algorithm was tested successfully for scenarios wherein local minima can often be encountered, like head-on collisions or passing through narrow pathways without the use of virtual forces or gyroscopic forces (see *Lee et al (2003)* and *chang et al (2003))*. This was achieved by dynamically changing the repulsive gain and the eccentricity of the elliptical contour of the potential field, but for scenarios involving a large number of moving obstacles (100-200) in a tight space, a significant jittering or oscillation was recorded. The jittering experienced in such a scenario can be improved by the dynamic step adjustment method (see *Sun et al. (2017))* and the performance will be improved in future studies.